\newif\ifJOURNAL
\JOURNALfalse
\newif\ifCONF
\CONFfalse
\newif\ifarXiv
\arXivfalse
\newif\ifWP
\WPfalse
\newif\ifFULL
\FULLfalse

\newif\ifLATIN
\LATINfalse

\arXivtrue

\ifarXiv\LATINtrue\fi	

\newif\ifnotJOURNAL	
\notJOURNALtrue
\ifJOURNAL\notJOURNALfalse\fi

\newif\ifnotarXiv	
\notarXivtrue
\ifarXiv\notarXivfalse\fi

\newif\ifTR		
\TRfalse
\ifarXiv\TRtrue\fi
\ifWP\TRtrue\fi
\newif\ifnotTR
\notTRtrue
\ifarXiv\notTRfalse\fi
\ifWP\notTRfalse\fi

\newif\ifnotLATIN	
\notLATINtrue
\ifLATIN\notLATINfalse\fi

\ifJOURNAL

\fi
\ifarXiv

  \newcommand{\GTPXVII}{GTP17arXiv}
\fi
\ifWP

  \newcommand{\GTPXVII}{GTP17}
\fi
\ifFULL

\fi

\ifnotLATIN
  \newcommand{\KolmogorovTikhomirov}{kolmogorov/tikhomirov:1959}
  \newcommand{\KolmogorovCRfull}{kolmogorov:1941CR}
  \newcommand{\KolmogorovStationary}{kolmogorov:1941}
\fi
\ifLATIN
  \newcommand{\KolmogorovTikhomirov}{kolmogorov/tikhomirov:1959latin}
  \newcommand{\KolmogorovCRfull}{kolmogorov:1941CR-latin}
  \newcommand{\KolmogorovStationary}{kolmogorov:1941-latin}
\fi

\ifJOURNAL
\documentclass[toc]{article}
\usepackage{amsmath,amsfonts,amssymb,latexsym,graphicx}
\newcommand{\Extra}[1]{}
\fi

\ifCONF
\documentclass[toc]{article}
\usepackage{amsmath,amsfonts,amssymb,latexsym,graphicx}
\newcommand{\Extra}[1]{}
\fi

\ifarXiv
\documentclass[toc]{article}
\usepackage{amsmath,amsfonts,amssymb,latexsym,graphicx}
\newcommand{\Extra}[1]{}
\fi

\ifWP
\documentclass[toc]{gtarticle}
\usepackage{amsmath,amsfonts,amssymb,latexsym,epsfig,graphicx}
\renewcommand{\Extra}[1]{#1}
\fi

\ifFULL
\documentclass{article}
\usepackage{amsmath,amsfonts,amssymb,latexsym,color,epigraph,graphicx,eepic}
\newcommand{\Extra}[1]{\red{#1}}
\newcommand{\red}[1]{\textcolor{red}{#1}}

\newcommand{\bluebegin}{\begingroup\color{blue}}
\newcommand{\blueend}{\endgroup}

\fi

\allowdisplaybreaks[4]

\newcommand{\Vladimir}{Vladimir}
\newcommand{\DOT}{.}

\ifnotLATIN
\input{OT2enc.def}

\usepackage{CJK}
\fi

\newcommand{\st}{\mathrel{\!|\!}}

\newcommand{\D}{\,\mathrm{d}}
\newcommand{\dd}{\mathrm{d}}

\newcommand{\PPP}{\mathcal{P}}		
\newcommand{\SSS}{\mathcal{S}}		

\newcommand{\Int}{\mathop{\mathrm{Int}}\nolimits}

\newcommand{\bbbr}{\mathbb{R}}		

\newtheorem{lemma}{Lemma}

\newtheorem{theorem}{Theorem}
\newenvironment{proof}
  {\trivlist\item[\hskip\labelsep\textbf{Proof}]}
  {\endtrivlist}

\newenvironment{Proof}[1]
  {\trivlist\item[\hskip\labelsep\textbf{Proof #1\,}]}
  {\endtrivlist}
\newcommand{\boxforqed}{\rule{.3em}{1.5ex}}
\newcommand{\qedtext}{\unskip\nobreak\hfil
  \penalty50\hskip1em\null\nobreak\hfil\boxforqed
  \parfillskip=0pt\finalhyphendemerits=0\endgraf}

\newenvironment{remark*}
  {\trivlist\item[\hskip\labelsep{\bfseries Remark}]\relax}
  {\endtrivlist}

\ifJOURNAL
\title{Competing with stationary prediction strategies}
\author{Vladimir Vovk\\[5mm]
 Computer Learning Research Centre\\
  Department of Computer Science\\
  Royal Holloway, University of London,
  Egham, Surrey TW20 0EX, UK\\
  \texttt{vovk@cs.rhul.ac.uk}}
\fi

\ifCONF
\title{Competing with stationary prediction strategies}
\author{Vladimir Vovk\\[5mm]
 Computer Learning Research Centre\\
  Department of Computer Science\\
  Royal Holloway, University of London,
  Egham, Surrey TW20 0EX, UK\\
  \texttt{vovk@cs.rhul.ac.uk}}
\fi

\ifarXiv
\title{Competing with stationary prediction strategies}
\author{Vladimir Vovk\\
\texttt{vovk{\rm@}cs.rhul.ac.uk}\\
\texttt{http://vovk.net}}
\fi

\ifWP
\title{Competing with stationary prediction strategies}
\author{Vladimir Vovk}

\fi

\ifFULL
\title{Competing with stationary prediction strategies}
\author{Vladimir Vovk\\
\texttt{vovk{\rm@}cs.rhul.ac.uk}\\
\texttt{http://vovk.net}}
\fi

\begin{document}
\maketitle
\begin{abstract}
  In this paper we introduce the class of stationary prediction strategies
  and construct a prediction algorithm
  that asymptotically performs as well as the best continuous stationary strategy.
  We make mild compactness assumptions but no stochastic assumptions
  about the environment.
  In particular,
  no assumption of stationarity is made about the environment,
  and the stationarity of the considered strategies
  only means that they do not depend explicitly on time;
  we argue that it is natural to consider only stationary strategies
  even for highly non-stationary environments.
\end{abstract}

\section{Introduction}
\label{sec:introduction}

This paper belongs to the area of learning theory
that has been variously referred to as prediction with expert advice,
competitive on-line prediction,
prediction of individual sequences,
and universal on-line learning;
see \cite{cesabianchi/lugosi:2006} for a review.
There are many proof techniques known in this field;
this paper is based on Kalnishkan and Vyugin's Weak Aggregating Algorithm
\cite{kalnishkan/vyugin:2005},
but it is possible that some of the numerous other techniques
could be used instead.

In Section \ref{sec:results} we give the main definitions
and state our main results, Theorems \ref{thm:deterministic-compact}--\ref{thm:randomized};
their proofs are given
in Sections \ref{sec:proof-deterministic-compact}--\ref{sec:proof-randomized}.
In Section \ref{sec:stationarity}
we informally discuss the notion of stationarity,
and Section \ref{sec:conclusion} concludes.

\section{Main results}
\label{sec:results}

The \emph{game of prediction} between Predictor and Reality
is played according to the following protocol
(of \emph{perfect information},
in the sense that either player can see the other player's moves made so far).

\bigskip

\noindent
\textsc{Prediction protocol}\nopagebreak
\begin{tabbing}
  \qquad\=\qquad\=\qquad\kill
  Reality announces $(\ldots,x_{-1},y_{-1},x_0,y_0)\in(\mathbf{X}\times\mathbf{Y})^{\infty}$.\\
  FOR $n=1,2,\dots$:\\
  \> Reality announces $x_n\in\mathbf{X}$.\\
  \> Predictor announces $\gamma_n\in\Gamma$.\\
  \> Reality announces $y_n\in\mathbf{Y}$.\\
  END FOR.
\end{tabbing}

\noindent
After Reality's first move the game proceeds in rounds numbered by the positive integers $n$.
At the beginning of each round $n=1,2,\ldots$ Predictor is given some signal $x_n$
relevant to predicting the following observation $y_n$.
The signal is taken from the \emph{signal space} $\mathbf{X}$
and the observations from the \emph{observation space} $\mathbf{Y}$.
Predictor then announces his prediction $\gamma_n$,
taken from the \emph{prediction space} $\Gamma$,
and the prediction's quality in light of the actual observation
is measured by a \emph{loss function}
$\lambda:\Gamma\times\mathbf{Y}\to\bbbr$.
At the beginning of the game Reality chooses the infinite past,
$(x_n,y_n)$ for all $n\le0$.

In the games of prediction traditionally considered in machine learning
there is no infinite past.
This situation is modeled in our framework by extending the signal space and observation space
by new elements ${?}\in\mathbf{X}$ and ${?}\in\mathbf{Y}$,
defining $\lambda(\gamma,{?})$ arbitrarily,
and making Reality announce the infinite past
$(\ldots,x_{-1},y_{-1},x_0,y_0)=(\ldots,{?},{?},{?},{?})$
and refrain from announcing $x_n={?}$ or $y_n={?}$ afterwards
(intuitively, $?$ corresponds to ``no feedback from Reality'').

We will always assume that the signal space $\mathbf{X}$,
the prediction space $\Gamma$,
and the observation space $\mathbf{Y}$
are non-empty topological spaces
and that the loss function $\lambda$ is continuous.
Moreover,
we are mainly interested in the case
where $\mathbf{X}$, $\Gamma$, and $\mathbf{Y}$ are locally compact metric spaces,
the prime examples being Euclidean spaces and their open and closed subsets.
Our first results will be stated for the case
where all three spaces $\mathbf{X}$, $\Gamma$, and $\mathbf{Y}$ are compact.

\begin{remark*}
  Our results can be easily extended to the case
  where the loss on the $n$th round is allowed to depend,
  in addition to $\gamma_n$ and $y_n$,
  on the past $\ldots,x_{n-1},y_{n-1},x_n$.
  This would, however, complicate the notation.
\end{remark*}

Predictor's strategies in the prediction protocol will be called
\emph{prediction strategies}
(or \emph{prediction algorithms},
when they are defined explicitly and we want to emphasize this).
Mathematically such a strategy is a function
$D:(\mathbf{X}\times\mathbf{Y})^{\infty}\times\mathbf{X}\times\{1,2,\ldots\}\to\Gamma$;
it maps each history $(\ldots,x_{n-1},y_{n-1},x_n)$
and the current time $n$ to the chosen prediction.
In this paper we will only be interested in continuous prediction strategies $D$
(according to the traditional point of view \cite{martin-lof:1970},
going back to Brouwer,
only continuous prediction strategies can be computable;
although it should be mentioned that nowadays
there are influential definitions of computability
\cite{blum/etal:1989,blum/etal:1998}
not requiring continuity).
An especially natural class of strategies
is formed by the \emph{stationary prediction strategies}
$D:(\mathbf{X}\times\mathbf{Y})^{\infty}\times\mathbf{X}\to\Gamma$,
which do not depend on time explicitly;
since the origin of time is usually chosen arbitrarily,
this appears a reasonable restriction
(see Section \ref{sec:stationarity} for a further discussion).

\subsection*{Universal prediction strategies: compact deterministic case}

In this and next subsections we will assume that the spaces $\mathbf{X},\Gamma,\mathbf{Y}$
are all compact.
A prediction strategy is \emph{CS universal} for a loss function $\lambda$ if
its predictions $\gamma_n$ satisfy
\begin{equation}\label{eq:dominates-deterministic-compact}
  \limsup_{N\to\infty}
  \Biggl(
    \frac1N
    \sum_{n=1}^N
    \lambda
    (\gamma_n,y_n)
    {}-
    \frac1N
    \sum_{n=1}^N
    \lambda
    \bigl(
      D(\ldots,x_{n-1},y_{n-1},x_n),y_n
    \bigr)
  \Biggr)
  \le
  0
\end{equation}
for any continuous stationary prediction strategy $D$
and any biinfinite $\ldots,x_{-1},y_{-1},x_{0},y_{0},x_{1},y_{1},\ldots$\,.
(``CS'' refers to the continuity and stationarity of the prediction strategies
we are competing with.)
\begin{theorem}\label{thm:deterministic-compact}
  Suppose $\mathbf{X}$ and $\mathbf{Y}$ are compact metric spaces,
  $\Gamma$ is a compact convex subset of a Banach space,
  and the loss function $\lambda(\gamma,y)$ is continuous in $(\gamma,y)$
  and convex in the variable $\gamma\in\Gamma$.
  There exists a CS universal prediction algorithm.
\end{theorem}
A CS universal prediction algorithm will be constructed in the next section.

\subsection*{Universal prediction strategies: compact randomized case}

When the loss function $\lambda(\gamma,y)$ is not convex in $\gamma$,
two difficulties appear:
\begin{itemize}
\item
  the conclusion of Theorem \ref{thm:deterministic-compact} becomes false
  if the convexity requirement is removed
  (\cite{kalnishkan/vyugin:2005}, Theorem 2);
\item
  in some cases the notion of a continuous prediction strategy becomes vacuous:
  e.g., there are no non-constant continuous stationary prediction strategies
  when $\Gamma=\{0,1\}$
  and $(\mathbf{X}\times\mathbf{Y})^{\infty}\times\mathbf{X}$ is connected
  (the latter condition is equivalent to $\mathbf{X}$ and $\mathbf{Y}$
  being connected---see \cite{engelking:1989}, Theorem 6.1.15).
\end{itemize}
To overcome these difficulties,
we consider randomized prediction strategies.
The proof of Theorem \ref{thm:deterministic-compact}
will give a universal, in a natural sense,
randomized prediction algorithm;
on the other hand,
there will be a vast supply of continuous stationary prediction strategies.

\begin{remark*}
  In fact,
  the second difficulty is more apparent than real:
  for example, in the binary case ($\mathbf{Y}=\{0,1\}$)
  there are many non-trivial continuous prediction strategies
  in the canonical form of the prediction game \cite{vovk:1990}
  with the prediction space redefined as the boundary of the set of superpredictions
  \cite{kalnishkan/vyugin:2005}.
\end{remark*}

A \emph{randomized prediction strategy} is a function
$D:(\mathbf{X}\times\mathbf{Y})^{\infty}\times\mathbf{X}\times\{1,2,\ldots\}\to\PPP(\Gamma)$
mapping the past complemented by the current time
to the probability measures on the prediction space;
$\PPP(\Gamma)$ is always equipped with the topology of weak convergence
(\cite{billingsley:1968};
this topology is also discussed, in the compact case,
in Section \ref{sec:proof-randomized-compact} below).
In other words, this is a prediction strategy
in the extended game of prediction with the prediction space $\PPP(\Gamma)$.
Analogously,
a \emph{stationary randomized prediction strategy} is a function
$D:(\mathbf{X}\times\mathbf{Y})^{\infty}\times\mathbf{X}\to\PPP(\Gamma)$.

Let us say that a randomized prediction strategy outputting $\gamma_n$
is \emph{CS universal} for a loss function $\lambda$ if,
for any continuous stationary randomized prediction strategy $D$
and any biinfinite $\ldots,x_{-1},y_{-1},x_{0},y_{0},x_{1},y_{1},\ldots$,
\begin{equation}\label{eq:dominates-randomized-compact}
  \limsup_{N\to\infty}
  \left(
    \frac1N
    \sum_{n=1}^N
    \lambda(g_{n},y_n)
    -
    \frac1N
    \sum_{n=1}^N
    \lambda(d_{n},y_n)
  \right)
  \le
  0
  \enspace
  \textrm{a.s.},
\end{equation}
where $g_{1},g_{2},\ldots,d_{1},d_{2},\ldots$ are independent random variables
distributed as
\begin{align}
  g_{n}
  &\sim
  \gamma_n\label{eq:distributed-1},\\
  d_{n}
  &\sim
  D(\ldots,x_{n-1},y_{n-1},x_n),\label{eq:distributed-2}
\end{align}
$n=1,2,\ldots$\,.
Intuitively,
the ``a.s.''\ in (\ref{eq:dominates-randomized-compact})
refers to the prediction strategies' internal randomization.
\begin{theorem}\label{thm:randomized-compact}
  Let $\mathbf{X}$, $\Gamma$, and $\mathbf{Y}$ be compact metric spaces
  and $\lambda$ be a continuous loss function.
  There exists a CS universal randomized prediction algorithm.
\end{theorem}

\subsection*{Simple reductions to the compact case}

In the following two subsections we will discuss the case
where the signal, prediction, and observation spaces
are not required to be compact.
The goal of this subsection is to show that the compact case
is not as special as it may seem,
as far as Theorem \ref{thm:randomized-compact} is concerned.
The rest of the paper does not depend on this subsection.

In general,
we might consider $\mathbf{X}$, $\Gamma$, and $\mathbf{Y}$
together with their fixed compactifications
$\overline{\mathbf{X}}$, $\overline{\Gamma}$, and $\overline{\mathbf{Y}}$
(without loss of generality we can and will assume that
$\mathbf{X}$, $\Gamma$, and $\mathbf{Y}$
are dense in their compactifications,
and then the compactifications will be the closures of the original spaces,
which explains our notation).
Let us suppose that $\lambda$ is bounded and continuous,
and, moreover, can be continuously extended to the product
$\overline{\Gamma}\times\overline{\mathbf{Y}}$
of the compactifications;
such an extension is then unique and will also be denoted $\lambda$.

If $\mathbf{X}$, $\Gamma$, and $\mathbf{Y}$
are Euclidean spaces their natural compactifications
might be chosen as Aleksandrov's one-point compactification
(\cite{engelking:1989}, Theorem 3.5.11),
the corresponding projective space
(with $\bbbr\mathrm{P}^L$ being the compactification of $\bbbr^L$),
or the corresponding closed unit ball
(with the interior of the closed unit ball in $\bbbr^L$
identified with $\bbbr^L$
by mapping a vector $v$ of length $l\in[0,1)$ in the former set
to the vector $(\tan(\pi l/2))v$).
The Stone--\v{C}ech compactification
(\cite{engelking:1989}, Section 3.6)
will usually be too large:
we will want our compactifications to be metrizable.

Theorem \ref{thm:randomized-compact} will remain true
if instead of assuming $\mathbf{X}$, $\Gamma$, and $\mathbf{Y}$ to be metric compacts
we assume that
$\overline{\mathbf{X}}$, $\overline{\Gamma}$, and $\overline{\mathbf{Y}}$
are metric compacts
and if in the definition of CS universality (\ref{eq:dominates-randomized-compact})
we only consider continuous stationary prediction strategies
that have a continuous extension to
$(\overline{\mathbf{X}}\times\overline{\mathbf{Y}})^{\infty}\times\overline{\mathbf{X}}$.

\begin{remark*}
  An elegant way to avoid considering compactifications
  would be to assume that $\mathbf{X}$, $\Gamma$, and $\mathbf{Y}$
  are metrizable proximity spaces
  (see \cite{engelking:1989}, Section 8.4, or \cite{naimpally/warrack:1970},
  where \cite{engelking:1989}'s ``proximity spaces'' are called ``separated proximity spaces'')
  and to consider only proximity prediction strategies.
  By Smirnov's theorem (\cite{engelking:1989}, Theorem 8.4.13 and also Theorem 8.4.9;
  \cite{naimpally/warrack:1970}, Theorem 7.7)
  a proximity space can be identified with the corresponding topological space
  equipped with a compactification.
  Assuming that the loss function $\lambda$ is a bounded proximity function,
  it can be uniquely continuously extended to the compactification
  $\overline{\Gamma}\times\overline{\mathbf{Y}}$
  (\cite{naimpally/warrack:1970}, Theorem 7.10),
  and every proximity stationary prediction strategy can be identified
  with a continuous function on the compactification
  $(\overline{\mathbf{X}}\times\overline{\mathbf{Y}})^{\infty}\times\overline{\mathbf{X}}$
  (by the same theorem).
  To ensure that the compactifications are metrizable,
  it is sufficient to assume that the proximity spaces are second-countable
  (i.e., have countable proximity weights;
  see \cite{naimpally/warrack:1970}, Theorem 8.14,
  and \cite{engelking:1989}, Theorem 4.2.8).
  We chose the slightly clumsier language of compactifications
  because the notion of a topological space is much more familiar
  than that of a proximity space.
\end{remark*}

\subsection*{Universal prediction strategies: deterministic case}

Let us say that a set in a topological space is \emph{precompact}
if its closure is compact.
In Euclidean spaces,
precompactness means boundedness.
In this and next subsections we drop the assumption of compactness
of $\mathbf{X}$, $\Gamma$, and $\mathbf{Y}$,
and so we have to redefine the notion of CS universality.

A prediction strategy outputting $\gamma_n\in\PPP(\Gamma)$
is \emph{CS universal}
for a loss function $\lambda$ if,
for any continuous stationary prediction strategy $D$
and for any biinfinite $\ldots,x_{-1},y_{-1},x_{0},y_{0},x_{1},y_{1},\ldots$,
\begin{multline}\label{eq:dominates-deterministic}
  \bigl(
    \{\ldots,x_{-1},x_0,x_1,\ldots\}
    \text{ and }
    \{\ldots,y_{-1},y_0,y_1,\ldots\}
    \text{ are precompact}
  \bigr)\\
  \Longrightarrow
  \limsup_{N\to\infty}
  \Biggl(
    \frac1N
    \sum_{n=1}^N
    \lambda(\gamma_n,y_n)
    -
    \frac1N
    \sum_{n=1}^N
    \lambda
    \bigl(
      D(\ldots,x_{n-1},y_{n-1},x_n),y_n
    \bigr)
  \Biggr)
  \le
  0.
\end{multline}
The intuition behind the antecedent of (\ref{eq:dominates-deterministic}),
in the Euclidean case,
is that the prediction algorithm
knows that $\left\|x_n\right\|$ and $\left\|y_n\right\|$ are bounded
but does not know an upper bound in advance.

Let us say that the loss function $\lambda$ is \emph{large at infinity}
if, for all $y^*\in\mathbf{Y}$,
\begin{equation*}
  \lim_{\substack{y\to y^*\\\gamma\to\infty}}
  \lambda(\gamma,y)
  =
  \infty
\end{equation*}
(in the sense that for each constant $M$
there exists a neighborhood $O_{y^*}\ni y^*$ and compact $C\subseteq\Gamma$ such that
$\lambda\left(\Gamma\setminus C,O_{y^*}\right)\subseteq(M,\infty)$).
Intuitively, we require that faraway $\gamma\in\Gamma$
should be poor predictions for nearby $y^*\in\mathbf{Y}$.
This assumption is satisfied for most of the usual loss functions
used in competitive on-line prediction.
\begin{theorem}\label{thm:deterministic}
  Suppose $\mathbf{X}$ and $\mathbf{Y}$ are locally compact metric spaces,
  $\Gamma$ is a convex subset of a Banach space,
  and the loss function $\lambda(\gamma,y)$ is continuous,
  large at infinity, and convex in the variable $\gamma\in\Gamma$.
  There exists a CS universal prediction algorithm.
\end{theorem}
To have a specific example in mind,
the reader might check that $\mathbf{X}=\bbbr^{K}$, $\Gamma=\mathbf{Y}=\bbbr^{L}$,
and $\lambda(\gamma,y):=\left\|y-\gamma\right\|$
satisfy the conditions of the theorem.

\subsection*{Universal prediction strategies: randomized case}

We say that a randomized prediction strategy
outputting randomized predictions $\gamma_n$
is \emph{CS universal} if,
for any continuous stationary randomized prediction strategy $D$
and for any biinfinite $\ldots,x_{-1},y_{-1},x_{0},y_{0},x_{1},y_{1},\ldots$,
\begin{multline}\label{eq:dominates-randomized}
  \bigl(
    \{\ldots,x_{-1},x_0,x_1,\ldots\}
    \text{ and }
    \{\ldots,y_{-1},y_0,y_1,\ldots\}
    \text{ are precompact}
  \bigr)\\
  \Longrightarrow
  \left(
    \limsup_{N\to\infty}
    \left(
      \frac1N
      \sum_{n=1}^N
      \lambda(g_{n},y_n)
      -
      \frac1N
      \sum_{n=1}^N
      \lambda(d_{n},y_n)
    \right)
    \le
    0
    \enspace
    \textrm{a.s.}
  \right),
\end{multline}
where $g_{1},g_{2},\ldots,d_{1},d_{2},\ldots$ are independent random variables
distributed according to (\ref{eq:distributed-1})--(\ref{eq:distributed-2}).
\begin{theorem}\label{thm:randomized}
  Let $\mathbf{X}$ and $\mathbf{Y}$ be locally compact metric spaces,
  $\Gamma$ be a metric space,
  and $\lambda$ be a continuous and large at infinity loss function.
  There exists a CS universal randomized prediction algorithm.
\end{theorem}

\section{Proof of Theorem \ref{thm:deterministic-compact}}
\label{sec:proof-deterministic-compact}

In the rest of the paper
we will be using the notation $\Sigma$ for $(\mathbf{X}\times\mathbf{Y})^{\infty}\times\mathbf{X}$.
By Tikhonov's theorem (\cite{engelking:1989}, Theorem 3.2.4)
this is a compact space;
it is also metrizable
(\cite{engelking:1989}, Theorem 4.2.2).
Another standard piece of notation throughout the rest of the paper
will be $\sigma_n:=(\ldots,x_{n-1},y_{n-1},x_n)\in\Sigma$.
Remember that $\lambda$, as a continuous function on a compact set,
is bounded below and above (\cite{engelking:1989}, Theorem 3.10.6).

Let $\Gamma^{\Sigma}$ be the set of all continuous functions
from $\Sigma$ to $\Gamma$
with the \emph{topology of uniform convergence},
generated by the metric
\begin{equation*}
  \hat\rho(D_1,D_2)
  :=
  \sup_{\sigma\in\Sigma}
  \rho
  \bigl(
    D_1(\sigma),D_2(\sigma)
  \bigr),
\end{equation*}
$\rho$ being the metric in $\Gamma$
(induced by the norm in the containing Banach space).
Since the topological space $\Gamma^{\Sigma}$ is separable
(\cite{engelking:1989}, Corollary 4.2.18
in combination with Theorem 4.2.8),
we can choose a dense sequence $D_1,D_2,\ldots$ in $\Gamma^{\Sigma}$.

\begin{remark*}
  The topology in $\Gamma^{\Sigma}$ is defined via a metric,
  and this is one the very few places in this paper where we need a specific metric
  (for brevity we often talk about ``metric spaces'',
  but this can always be replaced by ``metrizable topological spaces'').
  Without using the metric,
  we could say that the topology in $\Gamma^{\Sigma}$ is the compact-open topology
  (\cite{engelking:1989}, Section 3.4).
  Since $\Sigma$ is compact,
  the compact-open topology on $\Gamma^{\Sigma}$
  coincides with the topology of uniform convergence
  (\cite{engelking:1989}, Theorem 4.2.17).
  The separability of $\Gamma^{\Sigma}$ now follows
  from \cite{engelking:1989}, Theorem 3.4.16 in combination with Theorem 4.2.8.
\end{remark*}

The next step is to apply Kalnishkan and Vyugin's
\cite{kalnishkan/vyugin:2005}
Weak Aggregating Algorithm (WAA) to this sequence.
We cannot just refer to \cite{kalnishkan/vyugin:2005}
and will have to redo their derivation of the WAA's main property
since Kalnishkan and Vyugin only consider the case
of finitely many ``experts'' $D_k$
and finite $\mathbf{Y}$.
(Although in other respects
we will not need their algorithm in full generality
and so slightly simplify it.)

Let $q_1,q_2,\ldots$ be a sequence of positive numbers summing to 1,
$\sum_{k=1}^{\infty}q_k=1$.
Define
\begin{equation*}
  l_n^{(k)}
  :=
  \lambda
  \left(
    D_k(\sigma_n),y_n
  \right),
  \quad
  L_N^{(k)}
  :=
  \sum_{n=1}^N
  l_n^{(k)}
\end{equation*}
to be the instantaneous loss of the $k$th expert $D_k$ on the $n$th round
and his cumulative loss over the first $N$ rounds.
For all $n,k=1,2,\ldots$ define
\begin{equation*}
  w_n^{(k)}
  :=
  q_k
  \beta_n^{L_{n-1}^{(k)}},
  \quad
  \beta_n
  :=
  \exp
  \left(
    -\frac{1}{\sqrt{n}}
  \right)
\end{equation*}
($w_n^{(k)}$ are the weights of the experts to use on round $n$)
and
\begin{equation*}
  p_n^{(k)}
  :=
  \frac
  {w_n^{(k)}}
  {\sum_{k=1}^{\infty}w_n^{(k)}}
\end{equation*}
(the normalized weights;
it is obvious that the denominator is positive and finite).
The WAA's prediction on round $n$ is
\begin{equation}\label{eq:WAA}
  \gamma_n
  :=
  \sum_{k=1}^{\infty}
  p_n^{(k)}
  D_k(\sigma_n)
\end{equation}
(the series is convergent in the Banach space
since the compactness of $\Gamma$ implies
$\sup_{\gamma\in\Gamma}\left\|\gamma\right\|<\infty$,
and $\gamma_n\in\Gamma$ since
\begin{multline}\label{eq:convergence-to-0}
  \gamma_n
  -
  \sum_{k=1}^K
  \frac{p_n^{(k)}}{\sum_{k=1}^K p_n^{(k)}}
  D_k(\sigma_n)\\
  =
  \sum_{k=1}^K
  \left(
    1
    -
    \frac{1}{\sum_{k=1}^K p_n^{(k)}}
  \right)
  p_n^{(k)}
  D_k(\sigma_n)
  +
  \sum_{k=K+1}^{\infty}
  p_n^{(k)}
  D_k(\sigma_n)
  \to
  0
\end{multline}
as $K\to\infty$).

Let $l_n:=\lambda(\gamma_n,y_n)$ be the WAA's loss on round $n$
and
$
  L_N
  :=
  \sum_{n=1}^N
  l_n
$
be its cumulative loss over the first $N$ rounds.
\begin{lemma}[\cite{kalnishkan/vyugin:2005}, Lemma 9]\label{lem:9}
  The WAA guarantees that, for all $N$,
  \begin{equation}\label{eq:lemma9}
    L_N
    \le
    \sum_{n=1}^N
    \sum_{k=1}^{\infty}
    p_n^{(k)}
    l_n^{(k)}
    -
    \sum_{n=1}^N
    \log_{\beta_n}
    \sum_{k=1}^{\infty}
    p_n^{(k)}
    \beta_n^{l_n^{(k)}}
    +
    \log_{\beta_N}
    \sum_{k=1}^{\infty}
    q_k
    \beta_N^{L_N^{(k)}}.
  \end{equation}
\end{lemma}
The first two terms on the right-hand side of (\ref{eq:lemma9})
are sums over the first $N$ rounds of different kinds of mean of the experts' losses
(see, e.g., \cite{hardy/etal:1952}, Chapter III,
for a general definition of the mean);
we will see later that they nearly cancel each other out.
If those two terms are ignored,
the remaining part of (\ref{eq:lemma9}) is identical
(except that $\beta$ now depends on $n$)
to the main property of the ``Aggregating Algorithm''
(see, e.g., \cite{vovk:2001competitive}, Lemma 1).
All infinite series in (\ref{eq:lemma9}) are trivially convergent.
\begin{Proof}{of Lemma \ref{lem:9}}
  The proof is by induction on $N$.
  Assuming (\ref{eq:lemma9}),
  we obtain
  \begin{multline*}
    L_{N+1}
    =
    L_N + l_{N+1}
    \le
    L_N
    +
    \sum_{k=1}^{\infty}
    p_{N+1}^{(k)}
    l_{N+1}^{(k)}\\
    \le
    \sum_{n=1}^{N+1}
    \sum_{k=1}^{\infty}
    p_n^{(k)}
    l_n^{(k)}
    -
    \sum_{n=1}^N
    \log_{\beta_n}
    \sum_{k=1}^{\infty}
    p_n^{(k)}
    \beta_n^{l_n^{(k)}}
    +
    \log_{\beta_N}
    \sum_{k=1}^{\infty}
    q_k
    \beta_N^{L_N^{(k)}}
  \end{multline*}
  (the first ``$\le$'' used the ``countable convexity''
  $l_n\le\sum_{k=1}^{\infty}p_n^{(k)}l_n^{(k)}$,
  which follows from (\ref{eq:convergence-to-0}) and
  \begin{equation*}
    \lambda
    \left(
      \sum_{k=1}^K
      \frac{p_n^{(k)}}{\sum_{k=1}^K p_n^{(k)}}
      D_k(\sigma_n),
      y_n
    \right)
    \le
    \sum_{k=1}^K
    \frac{p_n^{(k)}}{\sum_{k=1}^K p_n^{(k)}}
    \lambda
    \left(
      D_k(\sigma_n),
      y_n
    \right)
  \end{equation*}
  if we let $K\to\infty$).
  Therefore,
  it remains to prove
  \begin{equation*}
    \log_{\beta_N}
    \sum_{k=1}^{\infty}
    q_k
    \beta_N^{L_N^{(k)}}
    \le
    -\log_{\beta_{N+1}}
    \sum_{k=1}^{\infty}
    p_{N+1}^{(k)}
    \beta_{N+1}^{l_{N+1}^{(k)}}
    +
    \log_{\beta_{N+1}}
    \sum_{k=1}^{\infty}
    q_k
    \beta_{N+1}^{L_{N+1}^{(k)}}.
  \end{equation*}
  By the definition of $p_n^{(k)}$
  this can be rewritten as
  \begin{equation*}
    \log_{\beta_N}
    \sum_{k=1}^{\infty}
    q_k
    \beta_N^{L_N^{(k)}}
    \le
    -\log_{\beta_{N+1}}
    \frac
    {
      \sum_{k=1}^{\infty}
      q_k
      \beta_{N+1}^{L_{N}^{(k)}}
      \beta_{N+1}^{l_{N+1}^{(k)}}
    }
    {
      \sum_{k=1}^{\infty}
      q_k
      \beta_{N+1}^{L_{N}^{(k)}}
    }
    +
    \log_{\beta_{N+1}}
    \sum_{k=1}^{\infty}
    q_k
    \beta_{N+1}^{L_{N+1}^{(k)}},
  \end{equation*}
  which after cancellation becomes
  \begin{equation}\label{eq:to-check}
    \log_{\beta_N}
    \sum_{k=1}^{\infty}
    q_k
    \beta_N^{L_N^{(k)}}
    \le
    \log_{\beta_{N+1}}
    \sum_{k=1}^{\infty}
    q_k
    \beta_{N+1}^{L_{N}^{(k)}}.
  \end{equation}
  The last inequality follows from the general result
  about comparison of different means
  (\cite{hardy/etal:1952}, Theorem 85),
  but we can also check it directly
  (following \cite{kalnishkan/vyugin:2005}).
  Let $\beta_{N+1}=\beta_N^a$,
  where $0<a<1$.
  Then (\ref{eq:to-check}) can be rewritten as
  \begin{equation*}
    \left(
      \sum_{k=1}^{\infty}
      q_k
      \beta_N^{L_N^{(k)}}
    \right)^a
    \ge
    \sum_{k=1}^{\infty}
    q_k
    \beta_{N}^{aL_{N}^{(k)}},
  \end{equation*}
  and the last inequality follows from the concavity of the function $t\mapsto t^a$.
  \qedtext
\end{Proof}

\begin{lemma}[\cite{kalnishkan/vyugin:2005}, Lemma 5]
  Let $L$ be an upper bound on $\left|\lambda\right|$.
  The WAA guarantees that, for all $N$ and $K$,
  \begin{equation}\label{eq:lemma5}
    L_N
    \le
    L_N^{(K)}
    +
    \left(
      L^2 e^L + \ln\frac{1}{q_K}
    \right)
    \sqrt{N}.
  \end{equation}
\end{lemma}
(There is no term $e^L$ in \cite{kalnishkan/vyugin:2005}
since it only considers non-negative loss functions.)
\begin{proof}
  From (\ref{eq:lemma9}),
  we obtain:
  \begin{align*}
    L_N
    &\le
    \sum_{n=1}^N
    \sum_{k=1}^{\infty}
    p_n^{(k)}
    l_n^{(k)}
    +
    \sum_{n=1}^N
    \sqrt{n}
    \ln
    \sum_{k=1}^{\infty}
    p_n^{(k)}
    \exp
    \left(
      -\frac{l_n^{(k)}}{\sqrt{n}}
    \right)
    +
    \log_{\beta_N}
    q_K
    +
    L_N^{(K)}\\
    &\le
    \sum_{n=1}^N
    \sum_{k=1}^{\infty}
    p_n^{(k)}
    l_n^{(k)}
    +
    \sum_{n=1}^N
    \sqrt{n}
    \left(
      \sum_{k=1}^{\infty}
      p_n^{(k)}
      \left(
        1
        -
        \frac{l_n^{(k)}}{\sqrt{n}}
        +
        \frac{\left(l_n^{(k)}\right)^2}{2n}
        e^L
      \right)
      -
      1
    \right)\\
    &\quad{}+
    \log_{\beta_N}
    q_K
    +
    L_N^{(K)}\\
    &=
    L_N^{(K)}
    +
    \frac12
    \sum_{n=1}^N
    \frac{1}{\sqrt{n}}
    \sum_{k=1}^{\infty}
    p_n^{(k)}
    \left(l_n^{(k)}\right)^2
    e^L
    +
    \sqrt{N}\ln\frac{1}{q_K}\\
    &\le
    L_N^{(K)}
    +
    \frac{L^2e^L}{2}
    \sum_{n=1}^N
    \frac{1}{\sqrt{n}}
    +
    \sqrt{N}\ln\frac{1}{q_K}
    \le
    L_N^{(K)}
    +
    \frac{L^2e^L}{2}
    \int_0^N
    \frac{\D t}{\sqrt{t}}
    +
    \sqrt{N}\ln\frac{1}{q_K}\\
    &\le
    L_N^{(K)}
    +
    L^2e^L\sqrt{N}
    +
    \sqrt{N}\ln\frac{1}{q_K}
  \end{align*}
  (in the second ``$\le$'' we used the inequalities $e^t\le1+t+\frac{t^2}{2}e^{\left|t\right|}$
  and $\ln t\le t-1$).
  \qedtext
\end{proof}

Now it is easy to prove Theorem \ref{thm:deterministic-compact}.
Let $\gamma_n$ be the predictions output by the WAA.
Consider any continuous stationary prediction strategy $D$.
Since every continuous function on a metric compact is uniformly continuous
(\cite{engelking:1989}, Theorem 4.3.32),
for any $\epsilon>0$ we can find $\delta>0$ such that
$\left|\lambda(\gamma_1,y)-\lambda(\gamma_2,y)\right|<\epsilon$
whenever $\rho(\gamma_1,\gamma_2)<\delta$.
We can further find $K$ such that $\hat\rho(D_K,D)<\delta$,
and (\ref{eq:lemma5}) then gives,
for all biinfinite $\ldots,x_{-1},y_{-1},x_{0},y_{0},x_{1},y_{1},\ldots$,
\begin{multline*}
  \limsup_{N\to\infty}
  \Biggl(
    \frac1N
    \sum_{n=1}^N
    \lambda(\gamma_n,y_n)
    -
    \frac1N
    \sum_{n=1}^N
    \lambda(D(\sigma_n),y_n)
  \Biggr)\\
  \le
  \limsup_{N\to\infty}
  \Biggl(
    \frac1N
    \sum_{n=1}^N
    \lambda(\gamma_n,y_n)
    -
    \frac1N
    \sum_{n=1}^N
    \lambda(D_K(\sigma_n),y_n)
  \Biggr)
  +
  \epsilon\\
  \le
  \limsup_{N\to\infty}
  \left(
    L^2e^L + \ln\frac{1}{q_K}
  \right)
  \frac{1}{\sqrt{N}}
  +
  \epsilon
  =
  \epsilon;
\end{multline*}
since $\epsilon$ can be arbitrarily small
the WAA is CS universal.

\section{Proof of Theorem \ref{thm:randomized-compact}}
\label{sec:proof-randomized-compact}

Let us first recall some useful facts about the probability measures
on a metric compact $\Omega$
(we will be following \cite{\GTPXVII}).
The Banach space of all continuous real-valued functions on $\Omega$
with the usual pointwise addition and scalar action
and the sup norm will be denoted $C(\Omega)$.
By one of the Riesz representation theorems
(\cite{dudley:2002}, 7.4.1; see also 7.1.1),
the mapping $\mu\mapsto I_{\mu}$,
where
$
  I_{\mu}(f):=\int_{\Omega}f\D\mu
$,
is a linear isometry
between the set of all finite Borel signed measures $\mu$ on $\Omega$
with the total variation norm
and the dual space $C'(\Omega)$ to $C(\Omega)$
with the standard dual norm
(\cite{rudin:1991}, Chapter 4).
We will identify the finite Borel signed measures $\mu$ on $\Omega$
with the corresponding $I_{\mu}\in C'(\Omega)$.
This makes the set $\PPP(\Omega)$ of probability measures on $\Omega$
a convex closed subset of $C'(\Omega)$.

We will be interested, however,
in a different topology on $C'(\Omega)$,
the weakest topology for which all evaluation functionals
$\mu\in C'(\Omega)\mapsto\mu(f)$, $f\in C(\Omega)$,
are continuous.
This topology is known as the \emph{weak${}^*$ topology}
(\cite{rudin:1991}, 3.14),
and the topology inherited by $\PPP(\Omega)$
is known as the \emph{topology of weak convergence}
(\cite{billingsley:1968}, Appendix III).
The point mass $\delta_{\omega}$, $\omega\in\Omega$,
is defined to be the probability measure concentrated at $\omega$,
$\delta_{\omega}(\{\omega\})=1$.
The simple example of a sequence of point masses $\delta_{\omega_n}$
such that $\omega_n\to\omega$ as $n\to\infty$ and $\omega_n\ne\omega$ for all $n$
shows that the topology of weak convergence is different from the dual norm topology:
$\delta_{\omega_n}\to\delta_{\omega}$ holds in one but does not hold in the other.

It is not difficult to check that $\PPP(\Omega)$ remains a closed subset of $C'(\Omega)$
in the weak${}^*$ topology
(\cite{bourbaki:integration}, III.2.7, Proposition 7).
By the Banach--Alaoglu theorem
(\cite{rudin:1991}, 3.15)
$\PPP(\Omega)$ is compact in the topology of weak convergence
(this is a special case of Prokhorov's theorem,
\cite{billingsley:1968}, Appendix III, Theorem 6).
In the rest of this paper,
$\PPP(\Omega)$
(and all other spaces of probability measures)
are always equipped with the topology of weak convergence.

Since $\Omega$ is a metric compact,
$\PPP(\Omega)$ is also metrizable
(by the well-known Prokhorov metric:
\cite{billingsley:1968}, Appendix III, Theorem 6).

Define
\begin{equation}\label{eq:expected-loss}
  \lambda(\gamma,y)
  :=
  \int_{\Gamma}
  \lambda(g,y)
  \gamma(\dd g),
\end{equation}
where $\gamma$ is a probability measure on $\Gamma$.
This is the loss function in a new game of prediction
with the prediction space $\PPP(\Gamma)$;
it is convex in $\gamma$.

Let us check that the loss function (\ref{eq:expected-loss}) is continuous.
If $\gamma_n\to\gamma$ and $y_n\to y$
for some $(\gamma,y)\in\PPP(\Gamma)\times\mathbf{Y}$,
\begin{equation*}
  \left|
    \lambda(\gamma_n,y_n)
    -
    \lambda(\gamma,y)
  \right|
  \le
  \left|
    \lambda(\gamma_n,y_n)
    -
    \lambda(\gamma_n,y)
  \right|
  +
  \left|
    \lambda(\gamma_n,y)
    -
    \lambda(\gamma,y)
  \right|
  \to
  0
\end{equation*}
(the first addend tends to zero because of the uniform continuity
of $\lambda:\Gamma\times\mathbf{Y}\to\bbbr$
and the second addend by the definition of the topology of weak convergence).

Unfortunately,
Theorem \ref{thm:deterministic-compact} cannot be applied
to the new game of prediction directly:
the theorem assumes that $\Gamma$ is a subset of a Banach space,
whereas the dual to an infinite-dimensional Banach space is never even metrizable
in the weak$^*$ topology
(\cite{rudin:1991}, 3.16).
The proof of Theorem \ref{thm:deterministic-compact}, however,
still works for the new game.

It is clear that the mixture (\ref{eq:WAA}) is a probability measure.
The result of the previous section is still true,
and the randomized prediction strategy (\ref{eq:WAA})
produces $\gamma_n\in\PPP(\Gamma)$ that are guaranteed to satisfy
\begin{equation}\label{eq:mean}
  \limsup_{N\to\infty}
  \left(
    \frac1N
    \sum_{n=1}^N
    \lambda(\gamma_n,y_n)
    -
    \frac1N
    \sum_{n=1}^N
    \lambda(D(\sigma_n),y_n)
  \right)
  \le
  0,
\end{equation}
for any continuous stationary randomized prediction strategy $D$.
The loss function is bounded in absolute value
by a constant $L$,
and so the law of the iterated logarithm
(see, e.g., \cite{shafer/vovk:2001}, (5.8))
implies that
\begin{align}
  \limsup_{N\to\infty}
  \frac
  {
    \left|
      \sum_{n=1}^N
      \bigl(
        \lambda(g_n,y_n)
        -
        \lambda(\gamma_n,y_n)
      \bigr)
    \right|
  }
  {
    \sqrt{2L^2N\ln\ln N}
  }
  &\le
  1,\label{eq:LIL-1}\\
  \limsup_{N\to\infty}
  \frac
  {
    \left|
      \sum_{n=1}^N
      \bigl(
        \lambda(d_n,y_n)
        -
        \lambda(D(\sigma_n),y_n)
      \bigr)
    \right|
  }
  {
    \sqrt{2L^2N\ln\ln N}
  }
  &\le
  1\label{eq:LIL-2}
\end{align}
with probability one.
Combining the last two inequalities with (\ref{eq:mean}) gives
\begin{equation*}
  \limsup_{N\to\infty}
  \left(
    \frac1N
    \sum_{n=1}^N
    \lambda(g_n,y_n)
    -
    \frac1N
    \sum_{n=1}^N
    \lambda(d_n,y_n)
  \right)
  \le
  0
  \enspace
  \textrm{a.s.}
\end{equation*}
Therefore, the WAA (applied to $D_1,D_2,\ldots$)
is a universal continuous randomized prediction strategy.

\section{Proof of Theorem \ref{thm:deterministic}}
\label{sec:proof-deterministic}

In view of Theorem \ref{thm:deterministic-compact},
we only need to get rid of the assumption of compactness
of $\mathbf{X}$, $\Gamma$, and $\mathbf{Y}$.

\subsection*{Game of removal}

The proofs of Theorems \ref{thm:deterministic} and \ref{thm:randomized}
will be based on the following game
(an abstract version of the ``doubling trick'',
\cite{cesabianchi/lugosi:2006})
played in a topological space $X$:

\bigskip

\noindent
\textsc{Game of removal $G(X)$}\nopagebreak
\begin{tabbing}
  \qquad\=\qquad\=\qquad\kill
  FOR $n=1,2,\dots$:\\
  \> Remover announces compact $K_n\subseteq X$.\\
  \> Evader announces $p_n\notin K_n$.\\
  END FOR.
\end{tabbing}
\textbf{Winner:}
Evader if the set $\left\{p_1,p_2,\ldots\right\}$ is precompact;
Remover otherwise.

\bigskip

\noindent
Intuitively,
the goal of Evader is to avoid being removed to the infinity.
Without loss of generality
we will assume that Remover always announces a non-decreasing sequence of compact sets:
$K_1\subseteq K_2\subseteq\cdots$.
\begin{lemma}[Gruenhage]\label{lem:Gruenhage}
  Remover has a winning strategy in $G(X)$
  if $X$ is a locally compact and paracompact space.
\end{lemma}
\begin{proof}
  We will follow the proof of Theorem 4.1 in \cite{gruenhage:2006}
  (the easy direction).
  If $X$ is locally compact and $\sigma$-compact,
  there exists a non-decreasing sequence $K_1\subseteq K_2\subseteq\cdots$
  of compact sets covering $X$,
  and each $K_n$ can be extended to compact $K^*_n$
  so that $\Int K^*_n\supseteq K_n$
  (\cite{engelking:1989}, Theorem 3.3.2).
  Remover will obviously win $G(X)$ choosing $K^*_1,K^*_2,\ldots$ as his moves.

  If $X$ is the sum of locally compact $\sigma$-compact spaces $X_s$, $s\in S$,
  Remover plays, for each $s\in S$, the strategy described in the previous paragraph
  on the subsequence of Evader's moves belonging to $X_s$.
  If Evader chooses $p_n\in X_s$ for infinitely many $X_s$,
  those $X_s$ will form an open cover of the closure of $\{p_1,p_2,\ldots\}$
  without a finite subcover.
  If $x_n$ are chosen from only finitely many $X_s$,
  there will be infinitely many $x_n$ chosen from some $X_s$,
  and the result of the previous paragraph can be applied.
  It remains to remember that each locally compact paracompact
  can be represented as the sum of locally compact $\sigma$-compact subsets
  (\cite{engelking:1989}, Theorem 5.1.27).
  \qedtext
\end{proof}

\subsection*{Large at infinity loss functions}

We will need the following useful property of large at infinity loss functions.
\begin{lemma}\label{lem:loss}
  Let $\lambda$ be a loss function that is large at infinity.
  For each compact set $B\subseteq\mathbf{Y}$ and each constant $M$
  there exists a compact set $C\subseteq\Gamma$ such that
  \begin{equation}\label{eq:loss}
    \forall\gamma\notin C,y\in B:
    \quad
    \lambda(\gamma,y)
    >
    M.
  \end{equation}
\end{lemma}
\begin{proof}
  For each point $y^*\in B$
  fix a neighborhood $O_{y^*}\ni y^*$
  and a compact set $C(y^*)\subseteq\Gamma$ such that
  $\lambda\left(\Gamma\setminus C(y^*),O_{y^*}\right)\subseteq(M,\infty)$.
  Since the sets $O_{y^*}$ form an open cover of $B$,
  we can find this cover's finite subcover
  $\{O_{y^*_1},\ldots,O_{y^*_n}\}$.
  It is clear that
  \begin{equation*}
    C
    :=
    \bigcup_{j=1,\ldots,n}
    C
    \left(
      O_{y^*_j}
    \right)
  \end{equation*}
  satisfies (\ref{eq:loss}).
  \qedtext
\end{proof}
In fact,
the only property of large at infinity loss functions that we will be using
is that in the conclusion of Lemma \ref{lem:loss}.
In particular, it implies the following lemma.
\begin{lemma}\label{lem:C-det}
  Under the conditions of Theorem \ref{thm:deterministic},
  for each compact set $B\subseteq\mathbf{Y}$
  there exists a compact convex set $C=C(B)\subseteq\Gamma$
  such that for each continuous stationary prediction strategy
  $D:\Sigma\to\Gamma$
  there exists a continuous stationary prediction strategy
  $D':\Sigma\to C$
  that dominates $D$ in the sense
  \begin{equation}\label{eq:prediction-type}
    \forall\sigma\in\Sigma,y\in B:
    \quad
    \lambda(D'(\sigma),y)
    \le
    \lambda(D(\sigma),y).
  \end{equation}
\end{lemma}
\begin{proof}
  Without loss of generality $B$ is assumed non-empty.
  Fix any $\gamma_0\in\Gamma$.
  Let
  \begin{equation*}
    M_1
    :=
    \sup_{y\in B}
    \lambda(\gamma_0,y),
  \end{equation*}
  let $C_1\subseteq\Gamma$ be a compact set such that  
  \begin{equation*}
    \forall \gamma\notin C_1,y\in B:
    \quad
    \lambda(\gamma,y)
    >
    M_1+1,
  \end{equation*}
  let
  \begin{equation*}
    M_2
    :=
    \sup_{(\gamma,y)\in C_1\times B}
    \lambda(\gamma,y),
  \end{equation*}
  and let $C_2\subseteq\Gamma$ be a compact set such that  
  \begin{equation*}
    \forall\gamma\notin C_2,y\in B:
    \quad
    \lambda(\gamma,y)
    >
    M_2+1.
  \end{equation*}
  It is obvious that $M_1\le M_2$ and $\gamma_0\in C_1\subseteq C_2$.
  We can and will assume $C_2$ convex
  (see \cite{rudin:1991}, Theorem 3.20(c)).

  Let us now check that $C_1$ lies inside the interior of $C_2$.
  Indeed, for any fixed $y\in B$ and $\gamma\in C_1$,
  we have $\lambda(\gamma,y)\le M_2$;
  since $\lambda(\gamma',y)>M_2+1$ for all $\gamma'\notin C_2$,
  some neighborhood of $\gamma$ will lie completely in $C_2$.

  Let $D:\Sigma\to\Gamma$
  be a continuous stationary prediction strategy.
  We will show that (\ref{eq:prediction-type}) holds
  for some continuous stationary prediction strategy $D'$
  taking values in the compact convex set $C(B):=C_2$.
  Namely,
  we define
  \begin{multline*}
    D'(\sigma)
    :=\\
    \begin{cases}
      D(\sigma) & \text{if $D(\sigma)\in C_1$}\\
      \frac{\rho(D(\sigma),\Gamma\setminus C_2)}{\rho(D(\sigma),C_1)+\rho(D(\sigma),\Gamma\setminus C_2)} D(\sigma)
      +\frac{\rho(D(\sigma),C_1)}{\rho(D(\sigma),C_1)+\rho(D(\sigma),\Gamma\setminus C_2)} \gamma_0
      & \text{if $D(\sigma)\in C_2\setminus C_1$}\\
      \gamma_0 & \text{if $D(\sigma)\in \Gamma\setminus C_2$}
    \end{cases}
  \end{multline*}
  where $\rho$ is the metric on $\Gamma$;
  the denominator $\rho(D(\sigma),C_1)+\rho(D(\sigma),\Gamma\setminus C_2)$
  is positive since already $\rho(D(\sigma),C_1)$ is positive.
  Since $C_2$ is convex,
  we can see that $D'$ indeed takes values in $C_2$.
  The only points $x$ at which the continuity of $D'$ is not obvious
  are those for which $D(\sigma)$ lies on the boundary of $C_1$:
  in this case
  one has to use the fact that $C_1$ is covered by the interior of $C_2$.

  It remains to check (\ref{eq:prediction-type});
  the only non-trivial case is $D(\sigma)\in C_2\setminus C_1$.
  By the convexity of $\lambda(\gamma,y)$ in $\gamma$,
  the inequality in (\ref{eq:prediction-type}) will follow from
  \begin{multline*}
    \frac{\rho(D(\sigma),\Gamma\setminus C_2)}{\rho(D(\sigma),C_1)+\rho(D(\sigma),\Gamma\setminus C_2)}
    \lambda(D(\sigma),y)\\
    +\frac{\rho(D(\sigma),C_1)}{\rho(D(\sigma),C_1)+\rho(D(\sigma),\Gamma\setminus C_2)}
    \lambda(\gamma_0,y)
    \le
    \lambda(D(\sigma),y),
  \end{multline*}
  i.e.,
  \begin{equation*}
    \lambda(\gamma_0,y)
    \le
    \lambda(D(\sigma),y).
  \end{equation*}
  Since the left-hand side of the last inequality is at most $M_1$
  and its right-hand side exceeds $M_1+1$,
  it holds true.
  \qedtext
\end{proof}
\begin{remark*}
  If the loss function is allowed to depend on the infinite past,
  the $\sigma$s in Lemma \ref{lem:C-det} will have to be restricted
  to a compact set $A\subseteq\Sigma$
  and the compact set $C$ will depend not only on $B$ but also on $A$
  (see Lemma 18 of \cite{\GTPXVII}).
\end{remark*}

\subsection*{The proof}

For each compact $B\subseteq\mathbf{Y}$
fix a compact convex $C(B)\subseteq\Gamma$ as in Lemma \ref{lem:C-det}.
Predictor's strategy ensuring (\ref{eq:dominates-deterministic})
is constructed from Remover's winning strategy in $G(\mathbf{X}\times\mathbf{Y})$
(see Lemma \ref{lem:Gruenhage};
metric spaces are paracompact by the Stone theorem,
\cite{engelking:1989}, Theorem 5.1.3)
and from Predictor's strategies $\SSS(A,B)$ outputting predictions
\begin{equation}\label{eq:gamma}
  \gamma_n\in C(B)
\end{equation}
and ensuring the consequent of (\ref{eq:dominates-deterministic})
for all continuous
\begin{equation}\label{eq:DABC}
  D:(A\times B)^{\infty}\times A\to C(B)
\end{equation}
under the assumption that $(x_n,y_n)\in A\times B$
for given compact $A\subseteq\mathbf{X}$ and $B\subseteq\mathbf{Y}$
(the existence of such $\SSS(A,B)$
is asserted in Theorem \ref{thm:deterministic-compact}).
Remover's moves are assumed to be of the form $A\times B$
for compact $A\subseteq\mathbf{X}$ and $B\subseteq\mathbf{Y}$.
Predictor is simultaneously playing the game of removal
$G(\mathbf{X}\times\mathbf{Y})$ as Evader.

At the beginning of the game of prediction
Predictor asks Remover to make his first move $A_1\times B_1$ in the game of removal;
without loss of generality
we assume that $A_1\times B_1$ contains all $(x_n,y_n)$, $n\le0$
(there is nothing to prove if $\{(x_n,y_n)\st n\le0\}$ is not precompact).
Predictor then plays the game of prediction using the strategy $\SSS(A_1,B_1)$
until Reality chooses $(x_n,y_n)\notin A_1\times B_1$
(forever if Reality never chooses such $(x_n,y_n)$).
As soon as such $(x_n,y_n)$ is chosen,
Predictor announces $(x_n,y_n)$ in the game of removal
and notes Remover's response $(A_2,B_2)$.
He then continues playing the game of prediction using the strategy $\SSS(A_2,B_2)$
until Reality chooses $(x_n,y_n)\notin A_2\times B_2$,
etc.

Let us check that this strategy for Predictor
will always ensure (\ref{eq:dominates-deterministic}).
If Reality chooses $(x_n,y_n)$ outside Predictor's current $A_k\times B_k$
finitely often,
the consequent of (\ref{eq:dominates-deterministic}) will be satisfied
for all continuous stationary $D:\Sigma\to C(B_K)$
($B_K$ being the second component of Remover's last move $(A_K,B_K)$)
and so, by Lemma \ref{lem:C-det},
for all continuous stationary $D:\Sigma\to\Gamma$.
If Reality chooses $(x_n,y_n)$ outside Predictor's current $A_k\times B_k$
infinitely often,
the set of $(x_n,y_n)$, $n=1,2,\ldots$, will not be precompact,
and so the antecedent of (\ref{eq:dominates-deterministic}) will be violated.

\section{Proof of Theorem \ref{thm:randomized}}
\label{sec:proof-randomized}

When $\gamma$ ranges over $\PPP(C)$
(identified with the subset of $\PPP(\Gamma)$
consisting of the measures concentrated on $C$)
for a compact $C\subseteq\Gamma$,
the loss function (\ref{eq:expected-loss}),
as we have seen, is continuous.
The following analogue of Lemma \ref{lem:C-det} will be useful.
\begin{lemma}\label{lem:C-rand}
  Under the conditions of Theorem \ref{thm:randomized},
  for each compact set $B\subseteq\mathbf{Y}$
  there exists a compact convex set $C=C(B)\subseteq\Gamma$
  such that for each continuous stationary randomized prediction strategy
  $D:\Sigma\to\PPP(\Gamma)$
  there exists a continuous stationary randomized prediction strategy
  $D':\Sigma\to\PPP(C)$
  such that (\ref{eq:prediction-type}) holds
  ($D'$ dominates $D$ ``on average'').
\end{lemma}
(In fact, this lemma is not needed
for the proof of Theorem \ref{thm:randomized} as we stated it,
but it will imply that $\gamma_n$ dominate $D(\sigma_n)$ on average,
for any continuous stationary randomized prediction strategy $D$:
see (\ref{eq:stage-K}).)
\begin{proof}
  Define $\gamma_0$, $M_1$, $C_1$, $M_2$, and $C_2$
  as in the proof of Lemma \ref{lem:C-det}.
  Fix a continuous function $f_1:\Gamma\to[0,1]$ such that $f_1=1$ on $C_1$
  and $f_1=0$ on $\Gamma\setminus C_2$
  (such an $f_1$ exists by the Tietze--Uryson theorem,
  \cite{engelking:1989}, Theorem 2.1.8).
  Set $f_2:=1-f_1$.
  Let $D:\Sigma\to\PPP(\Gamma)$ be a continuous stationary randomized prediction strategy.
  For each $\sigma\in\Sigma$,
  split $D(\sigma)$ into two measures on $\Gamma$
  absolutely continuous with respect to $D(\sigma)$:
  $D_1(\sigma)$ with Radon--Nikodym density $f_1$
  and $D_2(\sigma)$ with Radon--Nikodym density $f_2$;
  set
  \begin{equation*}
    D'(\sigma)
    :=
    D_1(\sigma)
    +
    \left|D_2(\sigma)\right|
    \delta_{\gamma_0}
  \end{equation*}
  (letting $\left|P\right|:=P(\Gamma)$ for $P$ a measure on $\Gamma$).
  It is clear that the stationary randomized prediction strategy $D'$ is continuous
  (in the topology of weak convergence, as usual),
  takes values in $\PPP(C_2)$,
  and
  \begin{multline*}
    \lambda(D'(\sigma),y)
    =
    \int_{\Gamma}
      \lambda(\gamma,y)
      f_1(\gamma)
    D(\sigma)(\dd\gamma)
    +
    \lambda(\gamma_0,y)
    \int_{\Gamma}
      f_2(\gamma)
    D(\sigma)(\dd\gamma)\\
    \le
    \int_{\Gamma}
      \lambda(\gamma,y)
      f_1(\gamma)
    D(\sigma)(\dd\gamma)
    +
    \int_{\Gamma}
      M_1
      f_2(\gamma)
    D(\sigma)(\dd\gamma)\\
    \le
    \int_{\Gamma}
      \lambda(\gamma,y)
      f_1(\gamma)
    D(\sigma)(\dd\gamma)
    +
    \int_{\Gamma}
      \lambda(\gamma,y)
      f_2(\gamma)
    D(\sigma)(\dd\gamma)
    =
    \lambda(D(\sigma),y)
  \end{multline*}
  for all $(\sigma,y)\in\Sigma\times B$.
  So we can take $C(B):=C_2$.
  \qedtext
\end{proof}
Fix one of the mappings $B\mapsto C(B)$
whose existence is asserted by the lemma.

We will prove that the prediction strategy of the previous section
with (\ref{eq:gamma}) replaced by
$
  \gamma_n\in\PPP(C(B))
$
and (\ref{eq:DABC}) replaced by
\begin{equation*}
  D:(A\times B)^{\infty}\times A\to\PPP(C(B))
\end{equation*}
is CS universal.
Let $D:\Sigma\to\PPP(\Gamma)$ be a continuous stationary randomized prediction strategy,
i.e., a continuous stationary prediction strategy
in the new game of prediction with loss function (\ref{eq:expected-loss}).
Let $(A_K,B_K)$ be Remover's last move
(if Remover makes infinitely many moves,
the antecedent of (\ref{eq:dominates-randomized}) is false,
and there is nothing to prove),
and let $D':\Sigma\to\PPP(C(B_K))$ be a continuous stationary randomized prediction strategy
satisfying (\ref{eq:prediction-type}) with $B:=B_K$.
From some $n$ on
our randomized prediction algorithm produces $\gamma_n\in\PPP(\Gamma)$
concentrated on $C(B_K)$,
and they will satisfy
\begin{multline}\label{eq:stage-K}
  \limsup_{N\to\infty}
  \left(
    \frac1N
    \sum_{n=1}^N
    \lambda(\gamma_n,y_n)
    -
    \frac1N
    \sum_{n=1}^N
    \lambda(D(\sigma_n),y_n)
  \right)\\
  \le
  \limsup_{N\to\infty}
  \left(
    \frac1N
    \sum_{n=1}^N
    \lambda(\gamma_n,y_n)
    -
    \frac1N
    \sum_{n=1}^N
    \lambda(D'(\sigma_n),y_n)
  \right)
  \le
  0.
\end{multline}
This is an interesting property
but slightly different from what Theorem \ref{thm:randomized} asserts.

According to the proof of Lemma \ref{lem:C-rand},
we can, and we will, assume that $D'(\sigma_n)$
generates outcomes $d'_n$ in two steps:
first $d_n$ is generated from $D(\sigma_n)$,
and then it is replaced by $\gamma_0$ with probability $f_2(\sigma_n)$.
The loss function is bounded in absolute value
on the compact set
$C(B_K)\times B_K$ by a constant $L$.
From the law of the iterated logarithm
(see (\ref{eq:LIL-1}) and (\ref{eq:LIL-2}))
applied to the losses of $\gamma_n$ and $d'_n$
we now obtain,
instead of (\ref{eq:stage-K}),
\begin{multline*}
  \limsup_{N\to\infty}
  \left(
    \frac1N
    \sum_{n=1}^N
    \lambda(g_n,y_n)
    -
    \frac1N
    \sum_{n=1}^N
    \lambda(d_n,y_n)
  \right)\\
  \le
  \limsup_{N\to\infty}
  \left(
    \frac1N
    \sum_{n=1}^N
    \lambda(g_n,y_n)
    -
    \frac1N
    \sum_{n=1}^N
    \lambda(d'_n,y_n)
  \right)\\
  =
  \limsup_{N\to\infty}
  \left(
    \frac1N
    \sum_{n=1}^N
    \lambda(\gamma_n,y_n)
    -
    \frac1N
    \sum_{n=1}^N
    \lambda(D'(\sigma_n),y_n)
  \right)
  \le
  0
  \enspace
  \textrm{a.s.};
\end{multline*}
it remains to compare this with (\ref{eq:dominates-randomized}).

\section{Stationarity and continuity}
\label{sec:stationarity}

As we said earlier,
the assumption of stationarity is very natural
for prediction strategies:
it just means that the arbitrary origin of time is not taken into account
(in the spirit of the invariance principle in statistics;
see, e.g., \cite{lehmann:1986}, Section 6.1).
Stationary strategies can detect and make use of all kinds of trends
and one-off phenomena;
e.g.,
they can perform well when the rate of environment change is constantly increasing
(as in our own environment).
There need not be stationarity in the environment.

Interestingly,
our prediction algorithms are continuous (or can be made continuous)
but not stationary.
First we discuss the continuity
of the prediction algorithms
constructed in the proofs of our four theorems.
\begin{description}
\item[Theorem \ref{thm:deterministic-compact}]
  It is easy to check that the WAA is continuous;
  by the Weierstrass $M$-test,
  (\ref{eq:WAA}) converges uniformly
  and so its sum is continuous.
\item[Theorem \ref{thm:randomized-compact}]
  To check that $\gamma_n$ is a continuous function of
  $\sigma_n$ in the topology of weak convergence,
  we only need to check that $\int f\D\gamma_n$ is a continuous function of $\sigma_n$
  for each $f\in C(\Sigma)$.
  This again follows from the Weierstrass $M$-test.
\item[Theorem \ref{thm:deterministic}]
  As described,
  Predictor's strategy is not continuous
  since his behavior changes suddenly when Reality outputs $(x_n,y_n)$
  outside his current $A_k\times B_k$,
  but it is clear that it can be ``smoothed around the edges''
  to ensure continuity.
\item[Theorem \ref{thm:randomized}]
  The situation is analogous to Theorem \ref{thm:deterministic}.
\end{description}

For concreteness,
we will discuss stationarity only in the case of Theorem \ref{thm:deterministic-compact}.
We know that the WAA is a prediction strategy that is continuous
as a function of the type $\Sigma\times\{1,2,\ldots\}\to\Gamma$.
It is not stationary
(i.e., we cannot get rid of the $\{1,2,\ldots\}$)
because it has to keep track of the experts' losses
since the beginning of the game of prediction.
Stationary strategies can depend on time only in a limited way:
e.g., in terms of our own environment,
they can depend on the time of day or the season.
But the WAA's dependence is much heavier:
it has to know precisely the time that has elapsed since the beginning.

Let us now check that
there are no universal continuous stationary prediction strategies
under conditions of Theorem \ref{thm:deterministic-compact}.
Suppose $\Gamma$ is such that there exists $f:\Gamma\to\Gamma$
without fixed points
(i.e., $f(\gamma)\ne\gamma$ for all $\gamma\in\Gamma$;
we can take, e.g., a circle as $\Gamma$).
If $D$ were a universal continuous stationary strategy,
we could define another continuous stationary strategy $D'(\sigma):=f(D(\sigma))$
and make Reality collude with $D'$
(i.e., output $y_n$ leading to a significantly smaller loss for $D'$;
this can be done for an appropriate choice of $\lambda$,
and in fact can be done for all usual $\lambda$).

\subsection*{Stationary Reality}

A standard problem in probability theory is where Reality
is governed by a stationary probability measure;
of course, only stationary prediction strategies are considered.
In this subsection we will list several references
for this problem,
considering, for simplicity, only the case where the signals $x_n$ are absent
(formally, we assume that $\mathbf{X}$ is a one-element set
and omit the $x_n$, which now do not carry any information, from our notation).

The problem of prediction has been studied extensively
for both strictly stationary sequences of observations
and wide sense stationary sequences
(the definitions and a general discussion of ``strict sense'' and ``wide sense'' concepts
can be found in \cite{doob:1953}, Chapter 2, Sections 8 and 3).
We will first assume that $\ldots,y_{-1},y_0,y_1,\ldots$
form a wide sense stationary sequence of random variables
and then a strictly stationary sequence.

The natural mode of prediction for wide sense stationary sequences
is linear prediction.
The problem of linear prediction
(not necessarily one-step-ahead, as in this paper)
of wide sense stationary sequences
was posed and solved by Kolmogorov
\cite{kolmogorov:1939,\KolmogorovCRfull,\KolmogorovStationary};
later but independently this was done by Wiener
\cite{wiener:1949}.

Kolmogorov and Wiener assumed the probability distribution of the observations known.
There are many efficient ways to estimate the spectral density of this probability distribution
(in terms of which the optimal linear predictor is expressed);
see, e.g., \cite{anderson:1971}, Chapter 9, for a review.
(An early idea of spectral estimation was proposed by Einstein in 1914:
see \cite{newton:2002}, p.~363.)

The problem of existence of universal prediction strategies
for strictly stationary and ergodic sequences of observations
was posed by Cover \cite{cover:1975},
and such strategies were found by Ornstein \cite{ornstein:1978}
for finite $\mathbf{Y}$
and Algoet \cite{algoet:1992} for $\mathbf{Y}$ a Polish space.
Papers \cite{gyorfi/etal:1999,gyorfi/lugosi:2001,nobel:2003}
construct such strategies
using techniques very similar to those of this paper.

\section{Conclusion}
\label{sec:conclusion}

An interesting direction of further research
is to obtain non-asymptotic versions of our results.
If the benchmark class of continuous stationary prediction strategies
is compact,
loss bounds can be given in terms of $\epsilon$-entropy
\cite{\KolmogorovTikhomirov}.
In general,
one can give loss bounds in terms of a nested family
of compact sets
whose union is dense in the set of continuous stationary prediction strategies
(in analogy with Vapnik and Chervonenkis's principle
of structural risk minimization \cite{vapnik:1998}).

\subsection*{Acknowledgments}

I am grateful to Yura Kalnishkan and Ilia Nouretdinov
for useful comments.
The construction of CS universal prediction strategies
is based on Alex Smola's and G\'abor Lugosi's suggestions.
This work was partially supported by MRC (grant S505/65).

\ifWP
  \DFlastpage
\fi
\end{document}